\documentclass[letterpaper, 10 pt, journal, twoside]{IEEEtran}

\usepackage{cite}
\usepackage{float}
\usepackage{multirow}
\usepackage{amssymb,amsfonts}
\usepackage{algorithmic}
\usepackage{booktabs}
\usepackage{graphicx}
\usepackage{textcomp}
\usepackage{xcolor}
\usepackage{algorithm}
\usepackage{mathrsfs}
\usepackage[caption=false,font=footnotesize]{subfig}
\usepackage{empheq}
\usepackage{comment}
\usepackage{mathtools, cuted}
\usepackage{bm}
\usepackage{dsfont}
\usepackage{xspace}

\usepackage{amsthm}
\usepackage[utf8]{inputenc}
\usepackage[english]{babel}
\usepackage[colorlinks,linkcolor=black,citecolor=black,urlcolor=black,bookmarks=false,hypertexnames=true]{hyperref} 
\usepackage[font=footnotesize,skip=4pt]{caption}
\usepackage{etoolbox}
\usepackage{soul}
\apptocmd{\thebibliography}{\scriptsize}{}{}

\usepackage{xcolor}
\definecolor{new_teal}{RGB}{0,110,120}      

\IEEEoverridecommandlockouts

\newtheorem*{definition*}{Definition}


\begin{document}
\title{GM-Loco: Terrain-Adaptive Humanoid Locomotion on Granular Media}

\author{Junnosuke Kamohara$^{1}$, 
        Feiyang Wu$^{1}$,
        Andy Ningan Zong$^{1}$, \\
        Daniel I. Goldman$^{2}$, 
        Yashwanth Nakka$^{1}$, 
        Seth Hutchinson$^{3}$, 
        Ye Zhao$^{1}$%
\thanks{$^{1}$Institute for Robotics and Intelligent Machines, Georgia Institute of Technology, Atlanta, GA 30332, USA 
        {\tt\footnotesize \{jkamohara3, nzong8, feiyangwu, cli895, ynakka3, yezhao\}@gatech.edu}
        }
\thanks{$^{2}$School of Physics, Georgia Institute of Technology, Atlanta, GA 30332, USA 
        {\tt\footnotesize daniel.goldman@physics.gatech.edu}
        }%
\thanks{$^{3}$Northeastern University, 360 Huntington Ave, Boston, MA 02115, USA 
        {\tt\footnotesize s.hutchinson@northeastern.edu}
        }%

\thanks{Digital Object Identifier (DOI): see top of this page.}
}


\IEEEaftertitletext{\vspace{-9\baselineskip}}
\maketitle

\begin{strip}
  \centering
  \IEEEaftertitletext{\vspace{-5.2\baselineskip}}
\end{strip}

\newcommand{\revision}[1]{\textcolor{teal}{\%#1}}
\newcommand*{\Revised}{\textcolor{black}}
\newcommand{\junnosuke}[1]{\textbf{\textcolor{blue}{#1}}}
\newcommand{\junnosukereply}[1]{\textbf{\textcolor{blue}{[Jn: #1]}}}
\newcommand{\feiyang}[1]{\textbf{\textcolor{purple}{Fy: #1}}}
\newcommand{\feiyangreply}[1]{\textbf{\textcolor{purple}{[Fy: #1]}}}

\newcommand{\ye}[1]{\textbf{\textcolor{red}{Ye: #1}}}
\newcommand{\Yez}[1]{{\color{gray}(ye: #1)}}
\newcommand{\seth}[1]{\textbf{\textcolor{green}{Seth: #1}}}
\newcommand{\nakka}[1]{\textbf{\textcolor{red}{Nakka: #1}}}


\newcommand{\resforce}{\boldsymbol{\alpha}}
\newcommand{\nvector}{\mathbf{e}_n}
\newcommand{\vvector}{\mathbf{e}_v}
\newcommand{\gvector}{\mathbf{e}_g}
\newcommand{\vecforcesimple}{\mathbf{G}}
\newcommand{\vecforce}{\mathbf{G}(\nvector, \vvector, \gvector)}
\newcommand{\nvecforce}{\mathbf{G}_n}
\newcommand{\tvecforce}{\mathbf{G}_t}
\newcommand{\gravity}{g}
\newcommand{\materialset}{\phi}
\newcommand{\density}{\rho_c}

\newcommand{\intfric}{\tilde{\mu}}
\newcommand{\surffric}{\mu}
\newcommand{\scalingmu}{f(\intfric)}

\newcommand{\mediascaling}{\xi}
\newcommand{\mediascalingmax}{\xi_\text{max}}
\newcommand{\mediascalingnorm}{\eta}
\newcommand{\mediascalingest}{\hat{\eta}}

\newcommand{\rvector}{\mathbf{e}_r}
\newcommand{\thetavector}{\mathbf{e}_\theta}
\newcommand{\zvector}{\mathbf{e}_z}
\newcommand{\normalprojection}{\mathbf{e}_{r\theta}}

\newcommand{\rvecforce}{\mathbf{G}_r}
\newcommand{\thetavecforce}{\mathbf{G}_\theta}
\newcommand{\zvecforce}{\mathbf{G}_z}

\newcommand{\selectionmat}{\mathbf{B}}
\newcommand{\jointpos}{\mathbf{q}_j}
\newcommand{\jointposinit}{\mathbf{q}_{0,j}}
\newcommand{\jointvel}{\dot{\mathbf{q}}_j}
\newcommand{\jointaccel}{\ddot{\mathbf{q}}_j}
\newcommand{\torque}{\boldsymbol{\tau}_j}
\newcommand{\massmat}{\mathbf{M}(\jointpos)}
\newcommand{\nonlinear}{\mathbf{b}(\jointpos, \jointvel)}
\newcommand{\contactjacobian}{\mathbf{J}_c(\jointpos)}
\newcommand{\contactwrench}{\mathbf{W}_c}
\newcommand{\contactforce}{\mathbf{F}_c}
\newcommand{\contactmoment}{\mathbf{M}_c}
\newcommand{\contactforcelement}{\mathbf{F}_{c,i}}
\newcommand{\jointnum}{N_j}
\newcommand{\numcontact}{N_c}
\newcommand{\anchorpoint}{\mathbf{p}_b}
\newcommand{\contactpoint}{\mathbf{p}_i}

\newcommand{\projgrav}{\mathbf{g}_b}
\newcommand{\projgravfeet}{\mathbf{g}_f}
\newcommand{\projgravxynorm}{\mathrm{\|\mathbf{g_b}\|_{xy}}}
\newcommand{\quatyaw}{\mathbf{q}_z}
\newcommand{\quatyawdes}{\mathbf{q}_z^*}
\newcommand{\linvel}{\mathbf{v}_{xy}}
\newcommand{\linveldes}{\mathbf{v}_{xy}^*}
\newcommand{\angvel}{\boldsymbol{\omega}_z}
\newcommand{\angveldes}{\boldsymbol{\omega}_z^*}
\newcommand{\zdes}{\mathrm{z}^*}
\newcommand{\vz}{\mathrm{v}_\mathrm{z}}
\newcommand{\wxy}{\boldsymbol{\omega}_{xy}}
\newcommand{\wxydot}{\boldsymbol{\dot{\omega}}_{xy}}
\newcommand{\action}{\mathbf{a}_t}
\newcommand{\prevaction}{\mathbf{a}_{t-1}}
\newcommand{\obs}{\mathbf{o}_t}
\newcommand{\obshist}{\mathbf{o}_{t-N:t}}
\newcommand{\reward}{r_t}
\newcommand{\privileged}{\mathbf{s}_t}
\newcommand{\latent}{\mathbf{z}_t}
\newcommand{\studentlatent}{\hat{\mathbf{z}}_t}
\newcommand{\studentaction}{\hat{\mathbf{a}}_t}
\newcommand{\footpos}{\mathbf{p}_f}
\newcommand{\footposleft}{\mathbf{p}_{f,L}}
\newcommand{\footposright}{\mathbf{p}_{f,R}}
\newcommand{\footvel}{\mathbf{v}_f}
\newcommand{\footcontact}{\mathbf{c}_f}
\newcommand{\footforce}{\mathbf{F}_c}
\newcommand{\tanfeetvel}{\|\dot{\mathbf{p}}_{\text{foot}, i}\|^2}
\newcommand{\leftfootypos}{p_{\text{f},0,y}}
\newcommand{\rightfootypos}{p_{\text{f},1,y}}
\newcommand{\stepwidth}{d}

\newcommand{\prior}{p(\mathbf{z}_t)}
\newcommand{\posterior}{p(\mathbf{z}_t | \mathbf{s}_t, \mathbf{o}_{t-N:t})}
\newcommand{\lossteacher}{L_\text{teacher}}
\newcommand{\lossppo}{L_\text{PPO}}
\newcommand{\lossvae}{L_\text{VAE}}
\newcommand{\lossstudent}{L_\text{student}}
\newcommand{\lossrepresentation}{L_\text{rep}}
\newcommand{\lossdistillation}{L_\text{dist}}
\newcommand{\piteacher}{\pi_\text{teacher}}
\newcommand{\pistudent}{\pi_\text{student}}
\newcommand{\phiteacher}{\phi_\text{teacher}}
\newcommand{\phistudent}{\phi_\text{student}}
\newcommand{\vaedecoder}{\psi}
\newcommand{\weightppo}{w_\text{PPO}}
\newcommand{\weightdist}{w_\text{dist}}

\newcommand{\sr}{\text{SR}\!\uparrow \mathrm{(\%)}}
\newcommand{\ev}{{e^v}\!\downarrow \mathrm{(m/s)}}
\newcommand{\eroll}{e^{\theta_x}\!\downarrow \mathrm{(deg)}}
\newcommand{\epitch}{e^{\theta_y}\!\downarrow \mathrm{(deg)}}
\newcommand{\eyaw}{e^{\theta}\!\downarrow \mathrm{(deg)}}
\newcommand{\projgravity}{\mathrm{\|\mathbf{g_b}\|_{xy}}\!\downarrow}
\newcommand{\angvelxy}{\mathrm{\|\wxy\|_2}\!\downarrow \mathrm{(rad/s)}}
\newcommand{\angaccelxy}{\mathrm{\|\wxydot\|_2}\!\downarrow \mathrm{(rad/s^2)}}
\newcommand{\dtw}{\text{DTW}\!\downarrow \mathrm{(m)}}
\newcommand{\pathtracking}{\text{Path Tracking}\!\downarrow \mathrm{(m)}}

\newcommand{\velerror}{e^v}
\newcommand{\yawerror}{e^{\theta}}

\newcommand{\srtext}{\text{success rate}\!\uparrow \mathrm{(\%)}}
\newcommand{\evtext}{\text{error vel.}\!\downarrow \mathrm{(m/s)}}
\newcommand{\eyawtext}{\text{error yaw}\!\downarrow \mathrm{(deg)}}
\newcommand{\projgravitytext}{\text{proj. gravity}\!\downarrow \mathrm{(-)}}

\newcommand{\cmark}{\ding{51}}
\newcommand{\xmark}{\ding{55}}

\newcommand{\ours}{Ours\xspace}
\newcommand{\baselineGMheuristic}{\textit{vanilla-GM-heuristic}\xspace}
\newcommand{\baselineGM}{\textit{vanilla-GM}\xspace}
\newcommand{\baselineRigid}{\textit{vanilla-rigid}\xspace}

\newcommand{\modelrigid}{Rigid\xspace}
\newcommand{\modeltwodrft}{Heuristic-RFT\xspace}
\newcommand{\modelconesingle}{Cone-RFT-Single\xspace}
\newcommand{\modelconemulti}{Cone-RFT-Multi\xspace}
\newcommand{\modelthreedrft}{3D-RFT\xspace}
\newcommand{\modelmpm}{MPM\xspace}

\newcommand{\valb}[1]{\textcolor{red!70!black}{\boldsymbol{#1}}}
\newcommand{\valstd}[2]{#1 \pm \textcolor{gray}{#2}}
\newcommand{\valstdb}[2]{\textcolor{red!70!black}{\boldsymbol{#1}} \pm \textcolor{gray}{#2}}

\begin{abstract}
Humanoid locomotion on granular terrain remains a significant challenge due to its complex foot-terrain interaction dynamics that are difficult to model.
Existing approaches either ignore granular contact dynamics or incorporate simplified normal force models with heuristic tangential components.
In this work, we present a physics-grounded granular contact model based on three-dimensional resistive force theory (3D RFT) and efficiently simulate granular terrain for reinforcement learning (RL) training.
Unlike traditional rigid contact models and simplified granular contact models with ad-hoc heuristics, our contact solver produces physically accurate granular intrusion dynamics without resorting to heuristics.
It captures realistic penetration and tangential drag during training, enabling the policy to learn behaviors that transfer reliably to real-world granular terrain where rigid contact models fail.
To adapt to varying terrain conditions, we train a terrain-adaptive locomotion controller via teacher-student RL, using a variational autoencoder to encode terrain information into a compact latent representation.
Simulation studies using material point method (MPM) with NVIDIA Newton demonstrate that our method generalizes to unseen granular terrains, achieves a significantly higher success rate than baselines, and demonstrates zero-shot terrain identification and adaptation.
We further validate our approach through extensive hardware experiments across diverse real-world granular terrains including basalt, dry sand, and beach sand.
To the best of our knowledge, this is the first demonstration of agile humanoid locomotion on real-world granular terrain. Project page: \url{https://humanoid-gm-locomotion.github.io/HUMANOID-GM/}
\end{abstract}

\vspace{-0.1in}

\section{Introduction}\label{sec:introduction}

\IEEEPARstart{R}{obust} and versatile locomotion 
is essential for legged systems to operate reliably across diverse real-world environments.
Among the most challenging settings for field deployment is deformable granular terrain, including sand, gravel, and loose soil, which are common in outdoor, unstructured environments.
Unlike rigid ground, granular media exhibit complex multi-phase behavior, acting as either a solid or a fluid depending on loading conditions.
This produces large, highly nonlinear contact forces that are difficult to model, making balancing and walking control substantially harder. 
Locomotion on granular terrain therefore remains a significant open problem for legged robots.
Existing approaches fall into two broad categories. Model-based methods incorporate terradynamics into stable gait planning~\cite{xiong2017stability, gosyne2018bipedial} or constraints in trajectory optimization~\cite{hubicki2016tractable, chang2020learning}, while learning-based methods train policies in simulation with domain randomization~\cite{miki2022learning, lee2020challenging, wu2025learn, radosavovic2024real} or explicit granular contact models~\cite{choi2023learning, luo2025mild}.

\begin{figure}[t]
    \centering
    \setlength{\tabcolsep}{0pt}%
    \hspace{0pt}%
    \begin{minipage}[t]{0.333\linewidth}
        \includegraphics[width=\linewidth]{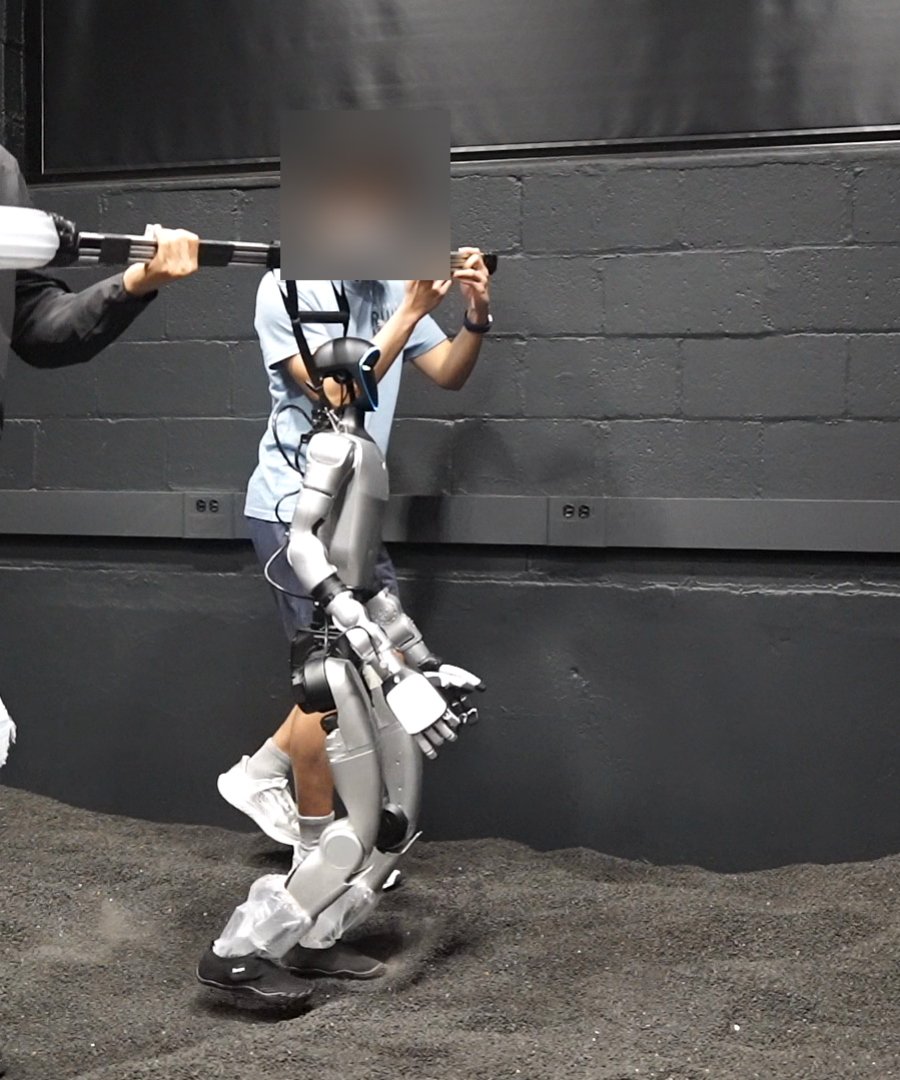}
    \end{minipage}
    \begin{minipage}[t]{0.333\linewidth}
        \includegraphics[width=\linewidth]{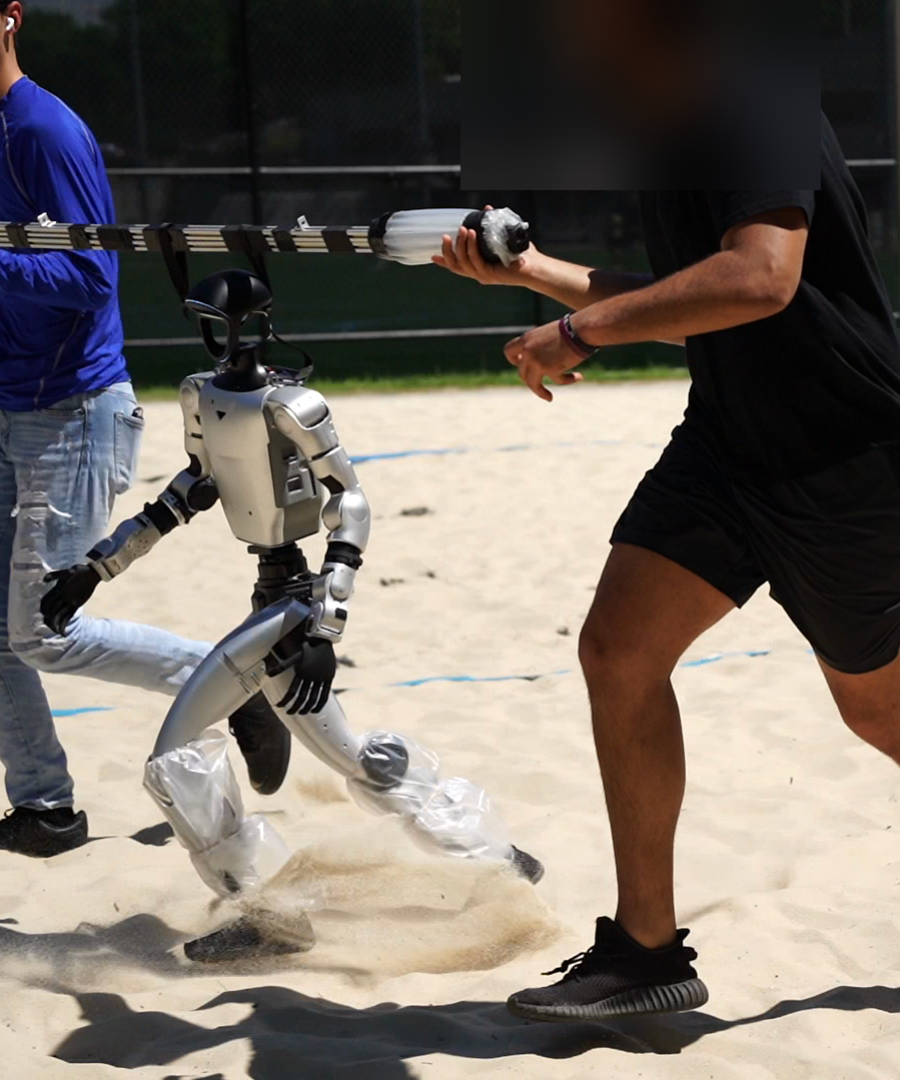}
    \end{minipage}
    \begin{minipage}[t]{0.333\linewidth}
        \includegraphics[width=\linewidth]{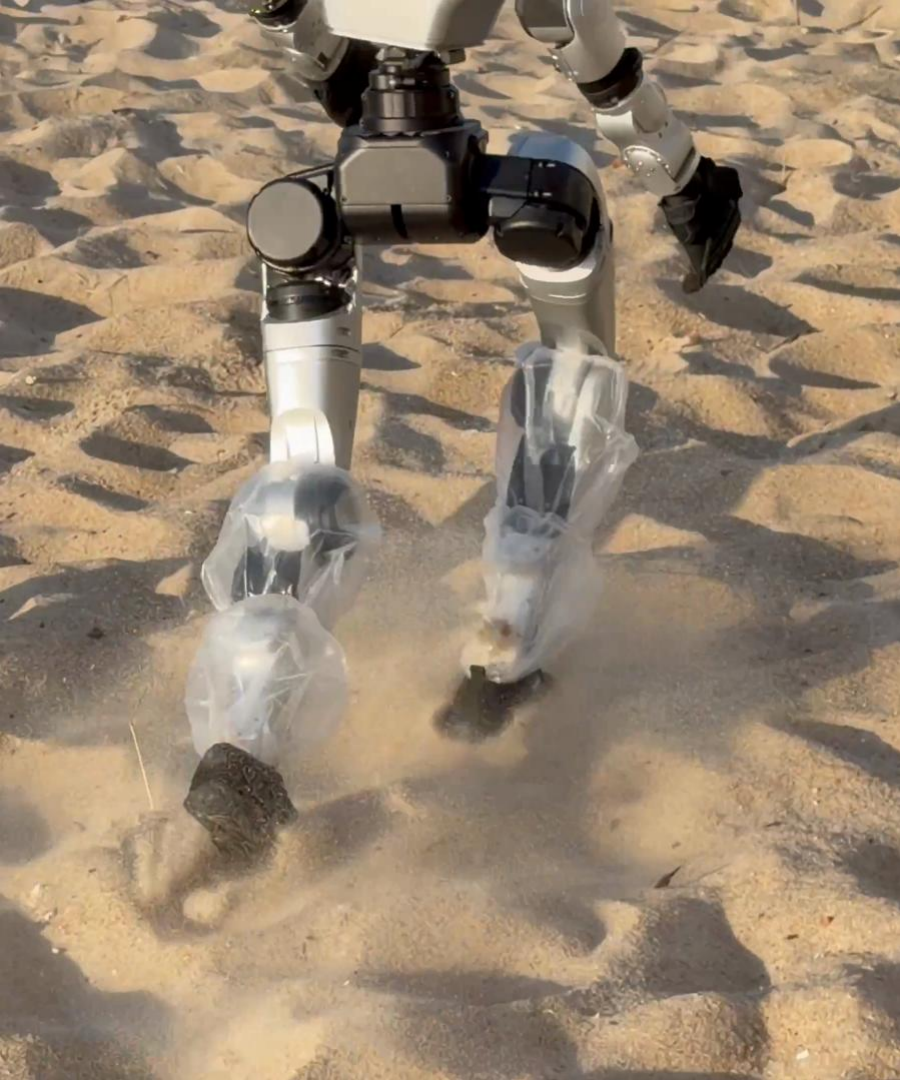}
    \end{minipage}
    \vspace{-3mm}
    \caption{
    The figure illustrates deployment of our policy on real-world granular terrains.
    Terrain corresponds to basalt (left), dry sand (center), and beach sand (right).
    }
    \label{fig:hardware-results}
    \vspace{-6mm}
\end{figure}

Model-based approaches tackle granular terrain through terradynamics, most notably resistive force theory (RFT)~\cite{li2013terradynamics, agarwal2021surprising, agarwal2023mechanistic} and its variants~\cite{chen2025sand}, which provide physically grounded models of leg and wheel intrusion in granular media.
These models have been embedded into the control of 1-D hoppers~\cite{hubicki2016tractable, chang2020learning} and bipedal robots~\cite{xiong2017stability, gosyne2018bipedial, chen2026bipedal}.
For hoppers, RFT has been integrated into trajectory optimization for impulsive jumping~\cite{hubicki2016tractable} and used to estimate ground reaction forces online~\cite{chang2020learning}.
For bipedal walkers, RFT provides stability criteria in hybrid zero dynamics~\cite{xiong2017stability} and zero-moment-point methods~\cite{gosyne2018bipedial}.
Chen \textit{et al.} estimate cost of transport from intrusion dynamics to evaluate the energy efficiency of bipedal locomotion~\cite{chen2026bipedal}.
However, these methods have primarily been validated on small-scale robots in simplified laboratory settings.
Their reliance on deterministic dynamics, predefined contact sequences, and limited modeling fidelity reduces robustness against the large perturbations typical of outdoor granular terrain, such as unexpected sinkage and foot slip.

Learning-based approaches, particularly reinforcement learning (RL), have shown strong potential for robust locomotion. 
A representative line of work achieves robust locomotion through domain randomization and privileged learning without explicitly modeling terrain dynamics~\cite{miki2022learning, lee2020challenging, wu2025learn, radosavovic2024real}.
These approaches do not capture granular contact behavior, relying instead on exhaustive domain randomization over rigid-contact dynamics.
To address this problem, recent work has incorporated granular contact models directly into the RL training environment~\cite{choi2023learning, luo2025mild}.
Prior work stitches together separate force model for normal-direction and tangential heuristics, whereas ours uses a single, experimentally validated 3D contact model with consistent physical grounding.
This yields a more physically grounded contact model that remains applicable across diverse real-world granular materials.

Real-world granular materials vary widely in mechanical properties, so a policy must adapt its motion to the terrain to remain robust.
Since terrain properties such as stiffness are not directly observable, we train the policy via teacher-student reinforcement learning with a learned terrain encoder.
The teacher policy has access to the ground-truth terrain stiffness, which the encoder compresses into a compact latent that conditions the policy.
The teacher is then distilled into a deployable student policy that infers the terrain latent from proprioceptive history alone, enabling the policy to adapt its motion to the inferred terrain conditions at deployment time.

In summary, our key contributions are:
\begin{itemize}
\item Integration of a three-dimensional granular contact model based on 3D RFT into the simulation pipeline.
\item A terrain-adaptive locomotion policy that produces adaptive swing foot clearance and gait frequency under changing terrain stiffness.
\item Demonstration of dynamic multi-gait humanoid locomotion, including circular walking, running, and jumping over simulation and real-world granular terrain.
\end{itemize}
\section{Related Work}

\subsection{Granular Contact Modeling}
Accurately modeling robot-foot-terrain interaction on granular media remains an open problem.
Mainstream rigid-body simulators~\cite{mujoco_warp, mittal2025isaac} assume non-penetrating contact, which fails to capture severe sinkage and slippage caused by granular terrain.
Particle-based methods such as the discrete element method (DEM)~\cite{zhang2024chrono} explicitly model inter-grain interactions and offer high simulation fidelity.
However, their computational cost prohibits use in online control or policy training, restricting them to small-scale offline analysis.
Continuum-based methods such as the material point method (MPM)~\cite{daviet2016semi} offer improved efficiency by discretizing the material on a spatial grid. 
Nevertheless, it still solves contact forces on a high-resolution spatial grid, which is computationally burdensome to scale for larger terrains and thousands of parallel simulations necessary for RL training.
To reduce computational cost, a variety of simplified contact models have been proposed, including spring-damper compliant models~\cite{leesoft}, linear viscoelastic models~\cite{lynch2020soft, lynch2024efficient}, and rate-dependent plastic models~\cite{ding2013foot, vasilopoulos2014compliant}.
However, they oversimplify the fluidic nature of granular mechanics and poorly align with real-world granular behavior.
In this work, we propose to incorporate Resistive force theory (RFT)~\cite{li2013terradynamics, agarwal2021surprising, treers2021granular, agarwal2023mechanistic} into the granular media contact model.
RFT is a force model for arbitrarily-shaped bodies that captures the resistive forces between an intruding body and granular media.
It computes the quasistatic reaction force on an intruding body as a function of the body's velocity angle, body orientation, 
and penetration depth.
Since RFT retains an analytical structure with parameters fit from intrusion experiments, it is a middle ground between simplified contact models and high-fidelity particle-based models.

\subsection{Legged Locomotion on Deformable Surface}
The robotics community has explored model-based and learning-based approaches to legged locomotion on granular terrain.
Model-based approaches incorporate granular contact models either as gait and contact force constraints or as system dynamics in trajectory optimization (TO)~\cite{xiong2017stability,gosyne2018bipedial,hubicki2016tractable,chang2020learning}.
For example, a few works define moment constraints by estimating the stability region of a linear inverted pendulum and imposing it in hybrid zero dynamics optimization~\cite{xiong2017stability}, while others synthesize stable gaits via zero moment point methods~\cite{gosyne2018bipedial}.
Other approaches embed terrain dynamics into TO, such as incorporating a terrain model into hopper dynamics for impulsive jumping on granular media~\cite{hubicki2016tractable} or learning ground reaction forces online via Gaussian processes~\cite{chang2020learning}.
Despite recent progress on locomotion over mud~\cite{godon2025mud} and dry sand~\cite{chen2025sand}, model-based methods generally suffer from model mismatch and remain primarily validated on low-dimensional robots in simplified laboratory settings.

Learning-based methods, particularly reinforcement learning (RL), offer a promising alternative.
A representative approach relies on domain randomization and privileged learning without explicitly modeling intricate terrain dynamics~\cite{miki2022learning, lee2020challenging, radosavovic2024real, wu2025learn}. 
These methods have demonstrated robust locomotion across natural terrain.
Fundamentally, these methods mitigate the model mismatch between a rigid contact model and granular terrain through exhaustive domain randomization, rather than principled deformable contact modeling.
To explicitly model granular contact dynamics, recent work incorporates granular media dynamics into the simulation pipeline during policy training.
Previous works utilize quasistatic RFT~\cite{li2013terradynamics} or a jammed granular cone model~\cite{choi2023learning} to describe granular intrusion dynamics in the normal direction for quadrupedal locomotion, and a similar approach applies to bipedal locomotion on a soft mattress~\cite{luo2025mild}.
However, these methods rely on a normal force model and supplement it with heuristic components such as Coulomb friction and spring-damper models for tangential forces, introducing ad-hoc design choices not grounded in granular contact physics. 
In this work, we develop a granular terrain contact model derived entirely from 3D RFT without resorting to heuristics, integrate it into IsaacLab~\cite{mittal2025isaac} to enable granular media simulation in a mainstream robotics simulator, and evaluate policy in Newton's MPM simulation~\cite{newton_physics,daviet2016semi}.
Furthermore, with this full 3D contact model, we focus on agile humanoid locomotion on granular terrain, including running and jumping, which remains largely unexplored in prior work.

\subsection{Terrain-Adaptive Locomotion}

Field deployment of humanoid robots often requires the robot to adjust foot clearance and gait cycle to remain robust under varying terrain conditions. 
Recent works have shown the effectiveness of latent variable representations for terrain-adaptive locomotion~\cite{kumar2021rma, lee2020challenging, luo2025mild}.
Other works estimate robot states~\cite{ji2022concurrent, choi2023learning, luo2025mild} and terrain properties~\cite{miki2022learning} as auxiliary tasks to improve locomotion robustness.
By concurrently training estimator networks that predict quantities such as foot clearance, contact probability, and base height, these methods adapt the policy to rough terrain.
Miki \textit{et al.} reconstruct terrain height and the friction coefficient as auxiliary tasks to enable terrain-adaptive policies~\cite{miki2022learning}.
The training pipeline follows a teacher-student RL framework that leverages an expert teacher to guide a student with a restricted observation space~\cite{chen2020learning,lee2020challenging,miki2022learning,wu2025learn}.
In this work, we adopt a terrain parameter estimation approach with teacher-student RL.
We concurrently train a teacher policy and a terrain encoder with proprioception and privileged information including ground truth robot states and terrain material properties.
Then, we distill the teacher policy into a student network that takes only proprioceptive history.
Unlike prior works, our approach recovers terrain stiffness from proprioception.
We demonstrate that this architecture enables terrain-adaptive behaviors including swing foot clearance modulation on soft granular surfaces.

\section{Preliminaries}
\label{sec:contact}

\begin{figure}[t]
    \noindent\begin{minipage}{\linewidth}
        \centering
        \includegraphics[width=1.0\linewidth]{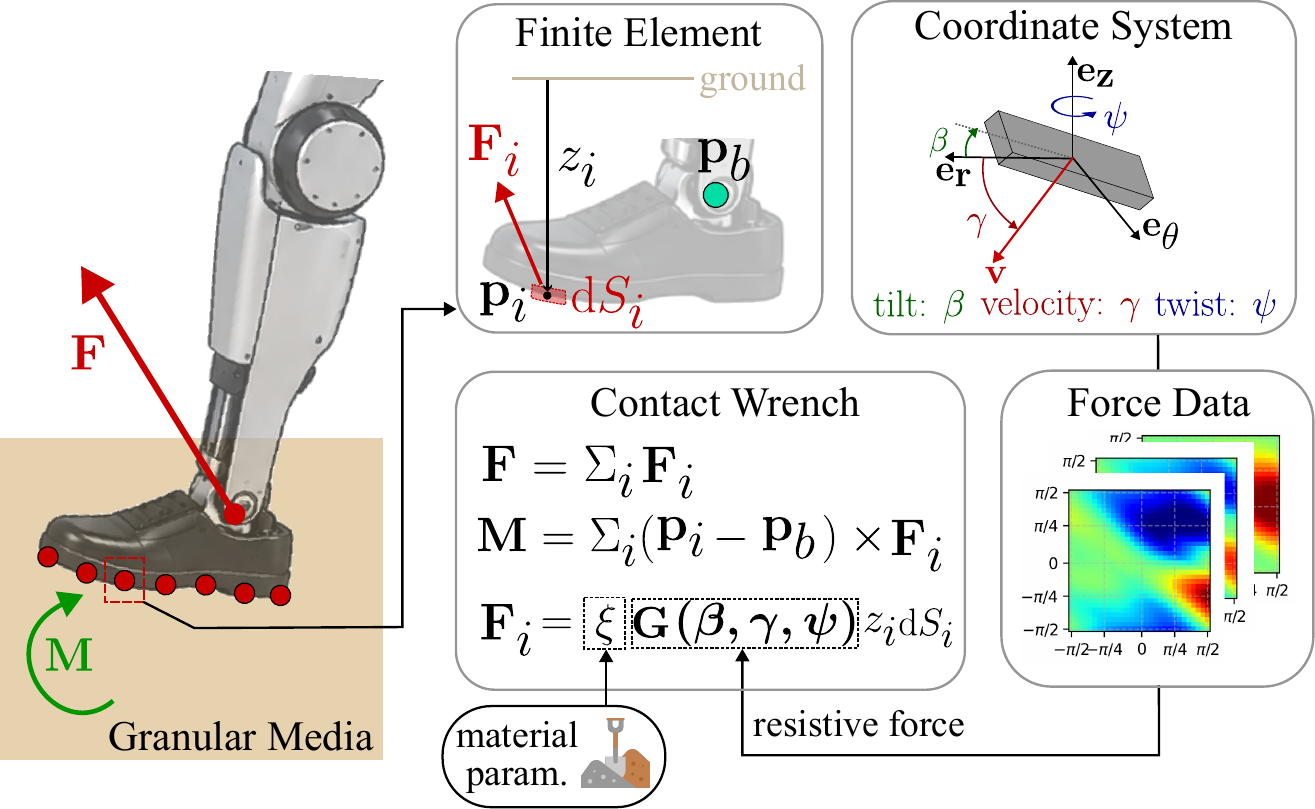}
    \end{minipage}
    \caption{
    Granular terrain contact model definition. 
    The terrain model predicts contact force $\mathbf{F}$ and moment $\mathbf{M}$ based on the intruder's penetration depth, velocity angle, and orientation.
    }
    \label{fig:rft-overview}
    \vspace{-3mm}
\end{figure}

We utilize 3D RFT~\cite{agarwal2023mechanistic} to model the contact forces between the humanoid foot and the granular terrain.
RFT assumes that the contact force on an intruding body is the sum of forces per area per depth $\resforce$ acting on infinitesimal elements: $\mathbf{F} = \int_S \resforce |z| \mathds{1}_{z<0} \mathrm{d} S$ where $z$ is the foot height relative to the terrain surface (negative when penetrating) and $S$ is the surface area, as demonstrated in Fig.~\ref{fig:rft-overview}.
Thus, we first derive the contact force and moment acting on each finite element.
Force per area per depth $\resforce$ (or resistive force) on a finite element is computed as follows:
\begin{equation}
    \resforce = \density g \scalingmu \vecforce
    \label{eq:1}
\end{equation}
where $\density$ is the effective material density, $\gravity$ is the gravitational constant, $\intfric$ is the internal coefficient of friction, $\nvector$ is the surface normal vector, $\vvector$ is the local velocity direction vector, and $\gvector$ is the gravity vector.
$\mediascaling=\density g \scalingmu$ represents the media-dependent scaling coefficient reflecting the intrusive strength of the medium, $\scalingmu$ is a scaling coefficient depending on the internal coefficient of friction $\intfric$, and $\vecforce$ is the generic RFT function expressed as $\vecforcesimple = c_1 \nvector + c_2 \vvector + c_3 \gvector$.
We use $\scalingmu = 894\intfric^3 - 386\intfric^2 + 89\intfric$ from the literature~\cite{agarwal2023mechanistic}.
Eq.~\ref{eq:1} is further modified to incorporate the friction cone as follows:
\begin{equation}
    \resforce = \density g \scalingmu \left[\nvecforce+\min\left(1, \frac{\surffric \| \nvecforce \|_2}{\| \tvecforce\|_2}\right)\tvecforce\right]
\end{equation}
where $\nvecforce$ and $\tvecforce$ are the normal and tangential components of the generic RFT function $\vecforcesimple$, and $\surffric$ is the coefficient of surface friction.
Therefore, the resistive force per area per depth $\resforce$ on each infinitesimal surface element is fully determined by $\vecforcesimple$.

We now re-express $\vecforcesimple$ in terms of an orthogonal basis and angles measured from these directions to align our formulation with the well-known 2D RFT~\cite{li2013terradynamics}.
To this end, we introduce an orthogonal basis $( \rvector, \thetavector, \zvector )$ and corresponding rotation angles $( \beta, \gamma, \psi )$ where $\beta$ is the polar angle between the surface normal $\nvector$ and the $z$-axis, $\gamma$ is the angle between the velocity direction and the $r$-axis, and $\psi$ is the surface twist angle between the $r$-axis and the projection of the surface normal onto the $r\theta$-plane denoted as $\normalprojection$.
By geometric decomposition, the unit vectors $(\nvector, \vvector, \gvector)$ are expressed in the orthogonal basis $( \rvector, \thetavector, \zvector )$ as follows:
\begin{align}
    \nvector &= \sin\beta\cos\psi \rvector + \sin\beta\sin\psi \thetavector - \cos\beta \zvector \nonumber \\
    \vvector &= \cos\gamma \rvector - \sin\gamma \zvector \nonumber \\
    \gvector &= -\zvector
\end{align}
By taking the dot product of $\vecforcesimple$ and each basis vector, we can express the generic RFT function in the orthogonal basis as follows:
\begin{align}
    \vecforcesimple &= \rvecforce \rvector + \thetavecforce \thetavector + \zvecforce \zvector \nonumber \\
    \rvecforce(\beta, \gamma, \psi) &= c_1 \sin\beta\cos\psi + c_2\cos\gamma \nonumber \\
    \thetavecforce(\beta, \gamma, \psi) &= c_1\sin\beta\sin\psi \nonumber \\
    \zvecforce(\beta, \gamma, \psi) &= -c_1\cos\beta - c_2\sin\gamma - c_3
\end{align}
where the characteristic angles $( \beta, \gamma, \psi )$ are computed from the orientation and velocity direction of the finite element, and the scalar functions $( c_1, c_2, c_3 )$ that depend on the characteristic angles are obtained by a $3$rd-order polynomial function fitted to plate intrusion experiment data generated via material point method (MPM) simulation.
More details can be found in \cite{agarwal2023mechanistic}.

We now derive the total contact force and moment applied to a humanoid robot's foot and integrate them into the robot's whole-body dynamics.
The humanoid robot is a floating-base system that can be described by whole-body dynamics (WBD):
\begin{equation}
    \massmat \jointaccel + \nonlinear = \selectionmat \torque + \sum_{c}^{\numcontact}\contactjacobian^\top \contactwrench
    \label{eq:wbd}
\end{equation}
where $\massmat \in \mathbb{R}^{(\jointnum + 6) \times (\jointnum + 6)}$ is the mass matrix, $\nonlinear \in \mathbb{R}^{\jointnum + 6}$ is the nonlinear term associated with Coriolis and gravity, $\selectionmat \in \mathbb{R}^{(\jointnum + 6) \times \jointnum}$ is the selection matrix that maps actuated joint torques to generalized coordinates, $\torque \in \mathbb{R}^{\jointnum}$ is the applied joint torque, $\jointpos, \jointvel, \jointaccel \in \mathbb{R}^{\jointnum + 6}$ are the generalized coordinate, velocity, and acceleration, $\numcontact$ is the number of contact bodies, $\contactjacobian \in \mathbb{R}^{6 \times (\jointnum + 6)}$ is the contact Jacobian of contact body $c$, and $\contactwrench = [\contactforce^\top, \contactmoment^\top]^\top \in \mathbb{R}^6$ is the contact wrench and composed of the contact force $\contactforce$ and moment $\contactmoment$, which we derive in the remainder of this section.
While the previous work in~\cite{luo2025mild} treats all finite element points as contact points in WBD, which results in a high-dimensional, aggregated contact Jacobian matrix, we instead sum forces on each foot and apply only the net wrench to WBD to reduce simulation cost.

For each foot body $c$, we predefine a collider as a set of $N$ grid points $\{\contactpoint \mid i = 1, \ldots, N\}$ on the body surface $\mathcal{S}_c$.
A grid point is treated as an active contact point when its $z$ position is below the surface level ($z=0$).
Using the resistive force $\resforce$ derived above, the discretized contact force at each active contact point follows  Eq.~\ref{eq:1}, given by $\contactforcelement = \resforce_i |z_i| \mathds{1}_{z_i < 0} \mathrm{d}S_i$, evaluated at contact point $i$.
Finally, we accumulate the contact forces and moments over all active contact points as follows: $\contactforce = \sum_i \contactforcelement$, $\contactmoment = \sum_i (\contactpoint - \anchorpoint) \times \contactforcelement$ where $\anchorpoint$ is the anchor point of contact body $c$, which is the ankle joint position in our setup.

\section{Method}

This section details our terrain-adaptive RL framework, which builds on the granular contact model introduced above.

\subsection{Terrain-Adaptive Reinforcement Learning}
\label{sec:rl}


\begin{figure}[t]
    \centering
    \includegraphics[width=1.0\linewidth]{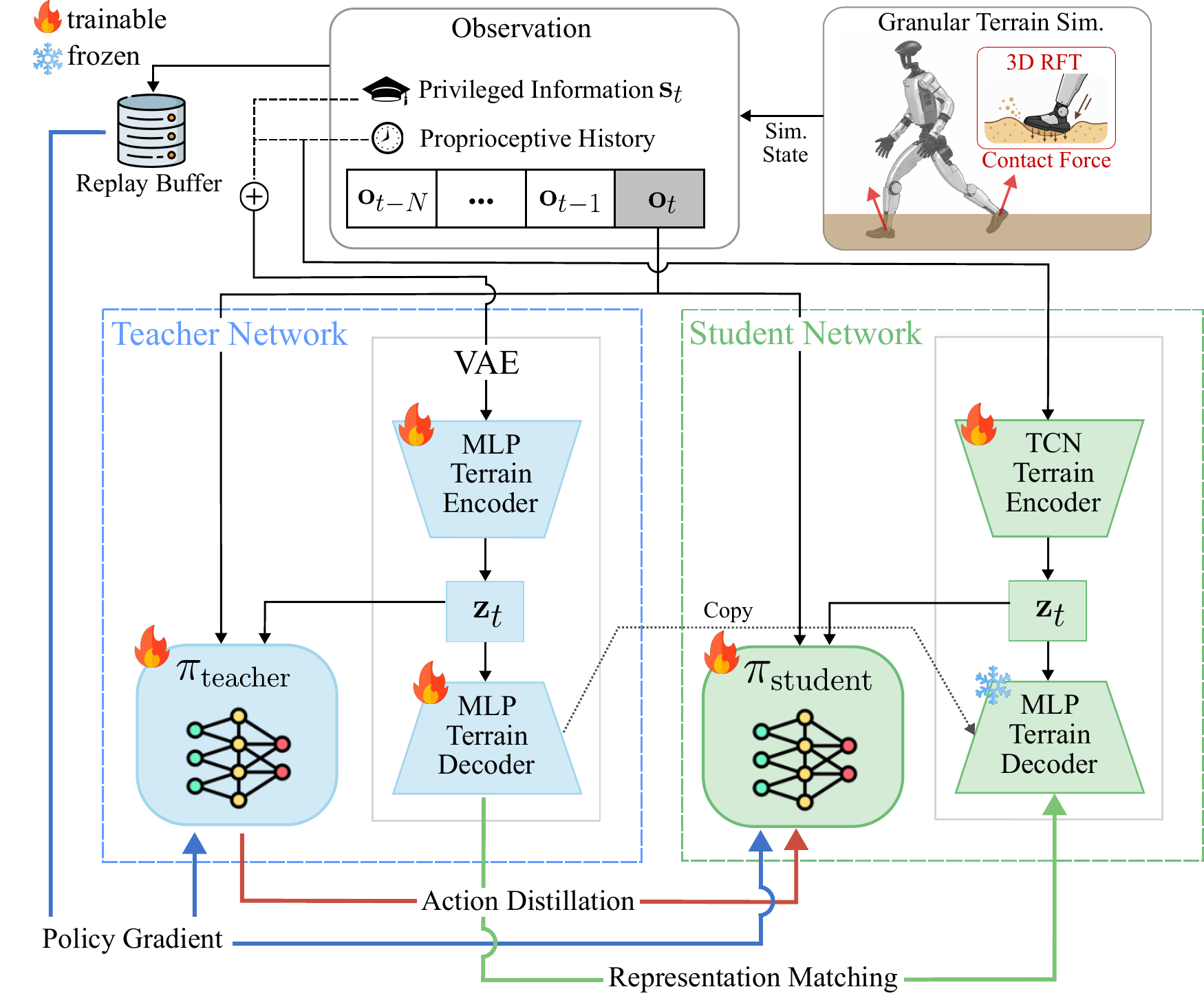}
    \caption{
    System diagram of our proposed terrain-adaptive humanoid locomotion.
    It has two-stage training where privileged teacher network is first trained with VAE terrain encoder and MLP policy.
    Then, this teacher policy is distilled to deployable student policy by teacher-student RL. 
    }
    \label{fig:system-overview}
    \vspace{-3mm}
\end{figure}

We formulate the velocity-tracking task for humanoid locomotion as a Markov Decision Process (MDP).
At each timestep $t$, the agent receives the observation $\mathbf{o}_t$, performs an action $\mathbf{a}_t$ according to its current policy $\pi(\cdot|\mathbf{o}_t)$, and obtains a scalar reward $R_t$. 
The goal of the agent is to maximize the discounted sum of rewards $\mathbb{E}_\pi [\sum_{t=0}^\infty \gamma^t R_t]$, where $\gamma\in [0,1)$ is the discount factor. 
Our control architecture leverages a terrain encoder as an adaptation module, trained via teacher-student RL.
Fig.~\ref{fig:system-overview} illustrates the overall architecture. 

\textbf{Teacher-Student RL:}
We leverage teacher-student RL~\cite{chen2020learning} where we first train the teacher policy $\piteacher(\action | \obs, \latent)$, which outputs the teacher action $\action$, using an asymmetric actor-critic method~\cite{pinto2017asymmetric} along with the terrain encoder.
In the asymmetric actor-critic setup, the teacher encoder and critic have access to privileged information that is not accessible on hardware, including ground truth robot states and terrain material properties. The teacher policy itself receives only the proprioceptive observation $\obs$ and the latent $\latent$.
The latent variable $\latent$ is produced by a learned terrain encoder from the privileged observation and proprioceptive history, encoding terrain information as detailed in the next paragraph.
We then distill the teacher policy and terrain encoder into a deployable student policy $\pistudent(\studentaction | \obs, \studentlatent)$, which outputs the student action $\studentaction$, and student encoder via behavior cloning of the teacher's policy actions and latent embeddings.

\textbf{Terrain Representation:}
We concurrently train a terrain encoder and decoder alongside the policy and condition the policy on the encoder's latent variable to capture terrain information.
The teacher encoder and decoder are trained jointly via a variational autoencoder (VAE) framework~\cite{kingma2013auto}, while the student encoder is trained by imitating the teacher encoder's output. The student has no decoder of its own and reuses the teacher's fixed decoder during training.
During teacher policy training, the teacher terrain encoder $\phiteacher(\latent | \privileged, \obshist)$ receives the privileged observation $\privileged$ and proprioceptive history $\obshist$ as input, and outputs the latent variable $\latent$ .
The resulting latent variable is concatenated with the current observation $\obs$ and processed by the actor network.
Unlike previous methods~\cite{lee2020challenging, kumar2021rma} which train the terrain encoder only via policy gradient loss, we leverage a variational auto encoder (VAE)~\cite{kingma2013auto} to learn a structured latent space, since its explicit reconstruction objective shapes the latent representation directly.
The decoder $\vaedecoder(\latent)$ is designed to reconstruct only the normalized media-dependent scaling coefficient $\mediascalingnorm = \vaedecoder(\latent) = \mediascaling / \mediascalingmax$, where $\mediascalingmax = \rho_\text{max} \, g \, \scalingmu|_{\intfric=\intfric_{\max}}$ is the maximum scaling coefficient across all terrain materials.
Reconstructing only $\mediascalingnorm$, rather than the full privileged state, steers the latent representation specifically toward terrain material properties.
During student policy training, the student terrain encoder $\phistudent(\obshist)$ receives the proprioceptive history $\obshist$ and outputs the latent variable $\studentlatent$.

\textbf{Loss Function:}
The teacher network is jointly optimized by the Proximal Policy Optimization (PPO)~\cite{schulman2017proximal} loss and the VAE loss, $\lossteacher = \lossppo + \lossvae$, with
\begin{equation}
    \lossvae = (\mediascalingnorm - \mediascalingest)^2 + \beta D_\text{KL}(\posterior \| \prior)
\end{equation}
where $\mediascalingest = \vaedecoder(\studentlatent)$ 
,
$\posterior$ is the posterior distribution of $\latent$ given the privileged observation $\privileged$ and proprioceptive history $\obshist$, and $\prior$ is the prior distribution of $\latent$ assumed to be $\mathcal{N}(\mathbf{0}, \mathbf{I})$.
We train the student network with a combined loss $\lossstudent = \weightppo \lossppo + \weightdist \lossdistillation + \lossrepresentation$, consisting of a PPO term $\lossppo$, an action distillation term $\lossdistillation = \|\action - \studentaction\|^2_2$, and a representation matching term.
\begin{equation}
    \lossrepresentation = \|\latent - \studentlatent\|^2_2 + (\vaedecoder(\latent) - \vaedecoder(\studentlatent))^2
\end{equation}

The distillation loss $\lossdistillation$ encourages the student to reproduce teacher actions, while the representation loss $\lossrepresentation$ enforces alignment between teacher and student encoders both in latent space and in decoded scaling coefficient space, encouraging the student encoder to identify terrain stiffness from proprioceptive history.
The weights $\weightppo$ and $\weightdist$ follow a curriculum inspired by previous work~\cite{radosavovic2024real}.

\subsection{Environment Design}
\label{sec:environment}

\textbf{Observation and Action Space:}
Proprioceptive information includes angular velocity $\angvel$, joint positions $\jointpos$, joint velocities $\jointvel$, and command velocity.
Additionally, we include the projected gravity $\projgrav$ derived from IMU data.
Finally, the actions from the previous timestep $\prevaction$ are also incorporated into the observation, which has been shown to produce smooth actions~\cite{rudin2022learning}. 
The privileged observation $\privileged$ includes base linear velocity $\linvel$, foot height in the global frame $\footpos$, foot contact state $\footcontact$, contact force $\contactforce$, current swing time since lift-off $\mathrm{T_s}$, and normalized media-dependent scaling coefficient $\mediascalingnorm$.
As with the teacher actor, the student actor also receives the proprioceptive observation $\obs$ and the latent representation $\studentlatent$ from the student terrain encoder that takes an $N$-step history of proprioceptive states $\obshist$ as input.
The action $\action \in \mathbb{R}^{29}$ represents the desired whole-body joint position offsets from a nominal pose $\jointposinit$, i.e., $\action = \jointpos^\text{des} - \jointposinit$.

\textbf{Reward Functions:}
Our reward design is summarized on our project page. 
We leverage existing reward functions implemented in IsaacLab for bipedal locomotion tasks, which include velocity command tracking, center of mass height tracking, and others.
Reward functions specific to our problem include penalties on the foot orientation, joint power, and contact impact.

\textbf{Implementation Detail:}
We employ PPO~\cite{schulman2017proximal}, an on-policy model-free RL algorithm, to train the teacher policy.
Both the actor and critic networks are implemented as separate three-layer MLPs with hidden sizes of ($512$, $256$, $128$) and ELU activations.
The teacher terrain encoder is a three-layer MLP with hidden sizes of ($256$, $128$, $64$) and ELU activations.
The decoder is a three-layer MLP with hidden sizes of ($64$, $128$, $256$) and ELU activations.
The student terrain encoder is implemented as a temporal convolutional network (TCN) with three one-dimensional convolutional layers followed by a single MLP. It receives an $N$-step history of proprioceptive states.
The policy is parameterized as a Gaussian distribution during training, with the mean and standard deviation predicted by the actor MLP.
To simulate deformable soft terrain, we implement the soft contact model with NVIDIA Warp~\cite{Macklin_Warp_A_High-performance_2022}, enabling simulation at hundreds of thousands of frames per second on a single RTX 4090 GPU.

\textbf{Training:}
We train the RL policy with $4096$ agents in parallel on IsaacLab~\cite{mittal2025isaac} using a single NVIDIA RTX $4090$ GPU.
For the teacher, we use PPO jointly with VAE and train it for $20,000$ iterations.
For the student, we use DAgger~\cite{ross2011reduction} with PPO and train it for $20,000$ iterations.
During both training and inference, RL runs at $50$\,Hz, while low-level PD control and physics simulation run at $200$\,Hz.

To handle real-world uncertainties associated with granular terrain and robot dynamics, we employ domain randomization with parameters listed on our project page. 
Furthermore, we introduce a training curriculum to facilitate stable policy learning.
We linearly ramp up the command velocity and tracking reward weight over the first $15,000$ iterations to introduce progressively challenging commands.
We also gradually lower the hard contact ground in the $-z$ direction as the robot successfully traverses terrain. 
This terrain curriculum ensures the robot first learns locomotion behaviors on hard terrain before transitioning to soft terrain.
\section{Results}
We evaluate the performance of the proposed control system for robust walking and running on different types of granular terrain in both simulation and real-world experiments.

\subsection{Simulation Evaluation}

We first evaluate the fidelity of our contact model in isolation through a single rigid body intrusion experiment, comparing horizontal travel distance, ground penetration, and ground reaction forces against the material point method (MPM).
We then examine how the contact model used during training affects the bipedal locomotion policy, evaluating torso stability and tracking accuracy under straight-line and turning walking, as well as robustness to unmodeled payload.

Next, we evaluate our terrain-adaptive policy across basalt, sand, and poppy seed terrain and compare it against policies trained with the rigid and 3D RFT contact models.
Finally, we analyze the policy's ability to perceive and adapt its behavior during transitions among terrains of varying stiffness.

\textbf{Simulation Setup: }
We test on high-fidelity granular terrain simulated with the MPM solver~\cite{daviet2016semi} from NVIDIA Newton physics~\cite{newton_physics}.
The evaluation uses two contact solvers: MuJoCo Warp~\cite{mujoco_warp} 
handles the robot's whole-body and rigid contact dynamics, while the MPM solver computes contact forces on deformable sand.
We use success rate (SR), velocity tracking error $\velerror$, and yaw tracking error $\yawerror$ as metrics.
The success rate is measured by running $100$ episodes and computing the proportion of episodes that complete the entire course of terrain.
An episode is considered a failure if the robot violates the base height constraint ($p_z < 0.2\,\text{m}$), the orientation constraint ($|\theta_x| > \tfrac{\pi}{4}$ or $|\theta_y| > \tfrac{\pi}{4}$\,rad), or fails to complete the terrain within $10$\,s.
The tracking error is measured as the mean absolute error between the commanded and measured base velocity, computed over $10$\,s episodes.

\begin{figure}[t]
    \centering
    \begin{minipage}{1.0\linewidth}
        \centering
        \includegraphics[width=1.0\linewidth]{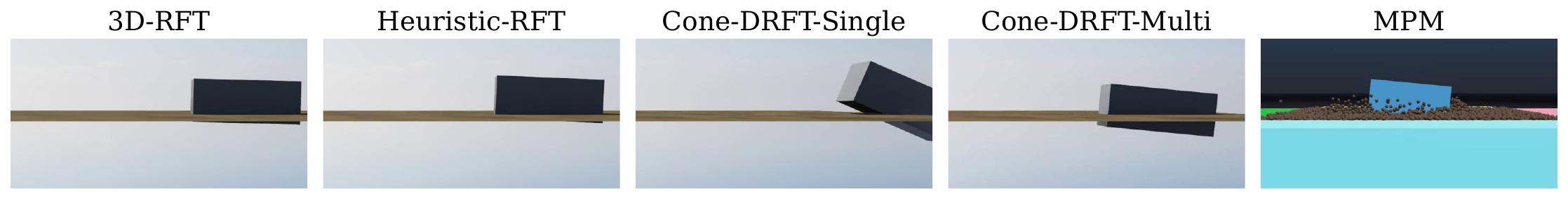}
    \end{minipage}
    \caption{
    Snapshots of a $35$\,kg box projected at $1.0$\,m/s under different contact models.
    From left to right : \modelthreedrft, \modeltwodrft~\cite{li2013terradynamics}, \modelconesingle~\cite{choi2023learning}, \modelconemulti~\cite{luo2025mild}, and \modelmpm~\cite{daviet2016semi}.
    }
    \label{fig:sim-contact-comparison}
    \vspace{-2mm}
\end{figure}

\textbf{3D RFT best captures three-dimensional granular intrusion dynamics.}
Next, we analyze the accuracy of our proposed contact model and its contribution to policy learning.
We first compare the horizontal travel distance and sinkage of a $35$\,kg object, whose mass matches the Unitree G1 robot's, under different contact models.
The object is initialized at a height of $0.1$\,m above the surface with a forward velocity of $1.0$\,m/s.
The compared models are (1) IsaacLab's default rigid contact model (\modelrigid), (2) 2D RFT~\cite{li2013terradynamics} with a Coulomb friction model (\modeltwodrft), (3) the jammed granular cone models~\cite{choi2023learning, luo2025mild} (\modelconesingle and \modelconemulti), (4) \modelthreedrft~\cite{agarwal2023mechanistic}, and (5) the material point method (\modelmpm)~\cite{daviet2016semi}.
We randomize the grain density $\density$ and the internal and surface friction coefficients $\intfric, \surffric$ across trials, and compare the resulting distributions of horizontal travel distance and sinkage.
Simulation results are shown in Fig.~\ref{fig:sim-contact-comparison}.
We treat Newton's MPM simulation as a continuum reference for granular intrusion.
We aim to evaluate which reduced-order models best reproduce the MPM's intrusion result.

\begin{figure}[t]
    \centering
    \includegraphics[width=1.0\linewidth]{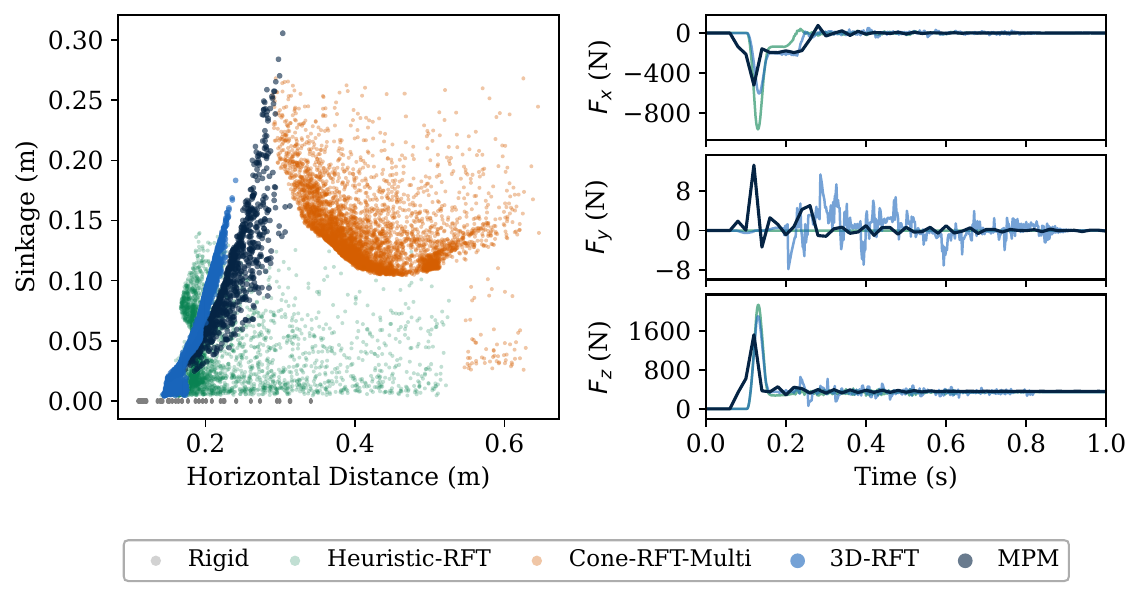}
    \caption{
    Horizontal travel distance and sinkage (left) and ground reaction force (GRF) profiles (right).
    On the left, each point corresponds to one trial with randomized material density, surface friction, and internal friction coefficients.
    }
    \label{fig:sim-contact-ablation}
    \vspace{-2mm}
\end{figure}

Fig.~\ref{fig:sim-contact-ablation} shows the statistics of the object's horizontal travel distance and sinkage (left) and the ground reaction force (GRF) profiles (right) under different contact models.
Across the randomized trials, our proposed contact model achieves the closest horizontal travel distance and sinkage to the MPM simulation, while \modeltwodrft and \modelconemulti produce a substantially larger horizontal travel distance.
\modelconesingle produces excessive slippage, as its single-point contact formulation does not hold for a bipedal robot with a large foot contact area.
It is therefore omitted from the main results.
The GRF in the $x$ and $z$ directions produced by our model also match the MPM simulation more closely than those of \modeltwodrft.
In the lateral direction, \modeltwodrft produces no force response, as its Coulomb-type tangential component acts only along the slip direction within the sagittal plane.
In contrast, the lateral force response of our model emerges naturally from the 3D RFT formulation rather than from a heuristically designed tangential model. 

\textbf{3D RFT facilitate the best RL policy performance.}
We also evaluate the contribution of each contact model to the RL policy performance of bipedal locomotion.
For this purpose, we train PPO policies with the contact models mentioned above except \modelconesingle  as it cannot provide stable simulation results to reliably train a policy.
We then evaluate each policy with MPM simulation on poppy seed using success rate and path tracking error, computed via dynamic time warping between the executed and commanded trajectories.
We select three evaluation scenarios, circular walking on flat terrain and rough terrain,  and straight walking on deep sand with a $30$\,cm depth and a low internal friction angle of $20^\circ$.
We generate rough terrain by initializing the z position of the MPM grid with a sinusoidal function.
More details on the terrain geometry are available on the supplementary project page.
These scenarios are chosen to highlight the performance difference among \modeltwodrft, \modelconemulti, and \modelthreedrft policies, especially when the terrain involves complex tangential-direction intrusion dynamics.
As shown in Table~\ref{tab:contact-ablation}, \modelthreedrft outperforms the baselines in all three scenarios, and the performance gap is more significant when the robot walks on rough terrain or deep sand.
This is because \modelthreedrft better captures tangential intrusion dynamics, which is crucial for stabilization in lateral direction.
While domain randomization helps improve the performance of \modeltwodrft and \modelconemulti policies, it is not sufficient to match the performance of \modelthreedrft under difficult terrain conditions.

\begin{table}[t]
    \centering
    \scriptsize
    \caption{
    Comparison among vanilla policies trained with different contact models.
    Policies are evaluated on poppy seed terrain simulated by MPM.
    The highest success rate and lowest errors are highlighted in bold.
    N/A indicates that the policy fails to achieve any successful episode; thus, tracking errors are not applicable.
    }
    \begin{minipage}{1.0\linewidth}
        \centering
        \begin{tabular}{c@{\hspace{4pt}}c@{\hspace{3pt}}c@{\hspace{3pt}}c@{\hspace{3pt}}}
            \toprule
            Task & Name &  $\sr$ & $\dtw$ \\
            \midrule
            \multirow{3}{*}{\shortstack{Circular walk\\flat}} & \modelthreedrft & $\valb{100}$ & $\valstdb{0.106}{0.024}$ \\
            {} & \modeltwodrft & $\valb{100}$ & $\valstd{0.142}{0.026}$  \\
            {} & \modelconemulti & $\valb{100}$ & $\valstd{0.173}{0.034}$ \\
            \midrule
            \multirow{3}{*}{\shortstack{Circular walk\\rough}} & \modelthreedrft & $\valb{100}$ & $\valstdb{0.141}{0.047}$ \\
            {} & \modeltwodrft & $\valb{100}$ & $\valstd{0.216}{0.042}$ \\
            {} & \modelconemulti & $\valb{100}$ & $\valstd{0.455}{0.101}$ \\
            \midrule
            \multirow{3}{*}{\shortstack{Walk\\deep sand}} & \modelthreedrft & $\valb{85}$ & $\valstdb{0.133}{0.077}$ \\
            {} & \modeltwodrft & $3$ & $\valstd{0.492}{0.229}$ \\
            {} & \modelconemulti & $8$ & $\valstd{0.353}{0.048}$ \\
            \bottomrule
        \end{tabular}%
    \end{minipage}
    \label{tab:contact-ablation}
\end{table}

\textbf{Our policy with 3D RFT achieves the best performance across diverse granular terrain.}
We assess the policy performance across basalt, sand, and poppy seed simulated by MPM. 
We vary the internal friction coefficient $\intfric$ and grain density $\density$ to respective values for each terrain.
Here, we compare three policies: (1) \modelrigid, (2) PPO policies trained with \modelthreedrft contact model, and (3) our proposed terrain-aware teacher-student policy (\ours) trained with 3D-RFT contact model.
We sample forward velocity commands from $1.0$ to $2.5$\,m/s to evaluate walking and running behaviors.
Results summarized in Table~\ref{tab:mpm-evaluation} demonstrate that the policies trained with \modelthreedrft achieve substantially higher success rates and tracking accuracy than \modelrigid, and \ours achieves the best tracking accuracy overall.
As evaluation terrain becomes softer, the success rate of \modelrigid drops sharply, confirming that accurate granular contact modeling during training is critical for locomotion over granular terrain.

\begin{table}[t]
    \centering
    \scriptsize
    \caption{
    Results of per-terrain evaluation with MPM simulation.
    We report success rate (SR), velocity tracking error ($e^v$), and yaw tracking error ($e^{\theta}$).
    The highest success rate and lowest tracking errors are highlighted in bold.
    N/A indicates that the policy fails to achieve any successful episode; thus, tracking errors are not applicable.
    }
    \begin{minipage}{1.0\linewidth}
        \centering
        \scriptsize
        \begin{tabular}{c@{\hspace{4pt}}c@{\hspace{4pt}}c@{\hspace{4pt}}c@{\hspace{5pt}}c@{\hspace{5pt}}c@{\hspace{5pt}}c}
        \toprule
        Terrain & Policy & Contact Model & $\sr$ & $\ev$ & $\eyaw$ \\
        \midrule
        \multirow{3}{*}{basalt} & \ours & \modelthreedrft & $\valb{100}$ & $\valstdb{0.29}{0.15}$ & $\valstdb{0.93}{0.27}$ \\
        {} & PPO & \modelthreedrft & $\valb{100}$ & $\valstd{0.35}{0.18}$  & $\valstd{2.71}{0.89}$ \\
        {} & PPO & \modelrigid & $98$ & $\valstd{0.35}{0.17}$ & $\valstd{1.55}{0.67}$ \\
        \midrule
        \multirow{3}{*}{sand} & \ours & \modelthreedrft & $\valb{100}$ & $\valstdb{0.33}{0.14}$  & $\valstdb{1.86}{0.47}$ \\
        {} & PPO & \modelthreedrft & $\valb{100}$  & $\valstd{0.35}{0.14}$  & $\valstd{2.14}{0.85}$ \\
        {} & PPO & \modelrigid & $0$ & N/A & N/A \\
        \midrule
        \multirow{3}{*}{poppy seed} & \ours & \modelthreedrft & $\valb{100}$ &  $\valstdb{0.35}{0.14}$ & $\valstd{2.07}{0.42}$ \\
        {} & PPO & \modelthreedrft & $\valb{100}$ &  $\valstd{0.36}{0.12}$ & $\valstdb{1.77}{0.73}$ \\
        {} & PPO & \modelrigid & $0$ & N/A & N/A \\
        \bottomrule
        \end{tabular}
    \end{minipage}
    \label{tab:mpm-evaluation}
\end{table}

\begin{figure}[t]
    \centering
    \begin{minipage}[t]{0.49\linewidth}
        \centering
        \includegraphics[width=\linewidth]{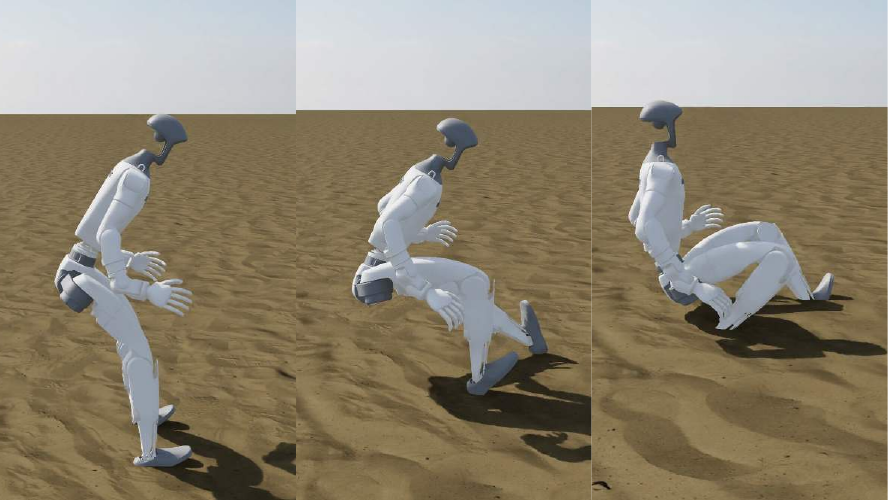}
    \end{minipage}\hfill
    \begin{minipage}[t]{0.49\linewidth}
        \centering
        \includegraphics[width=\linewidth]{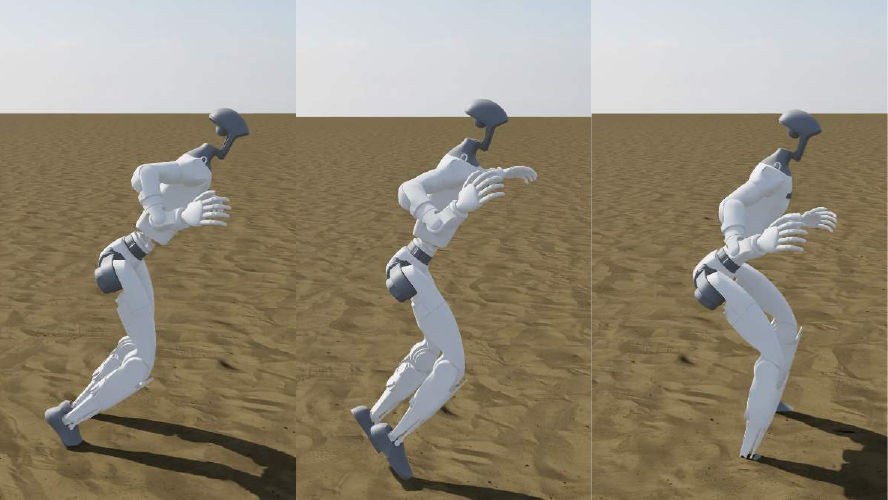}
    \end{minipage}
    \caption{
    Jumping motion achieved by PPO-Rigid (left) and PPO-3D-RFT (right). 
    PPO-Rigid fails to compensate ground slippage before flight phase, whereas PPO-3D-RFT successfully achieves the jump.
    }
    \label{fig:jumping}
    \vspace{-3mm}
\end{figure}

\textbf{3D RFT enables dynamic jumping on sand.}
Beyond bipedal walking, our method is applicable to dynamic jumping motion.
To this end, we train a jumping controller based on imitation RL and assistive wrench from ZEST~\cite{sleiman2026zest} using the \modelthreedrft contact model.
The policy is trained to imitate a kinematic joint trajectory retargeted from human motion capture data.
As demonstrated in Fig.~\ref{fig:jumping}, the policy successfully achieves jumping on simulated granular terrain.
Further implementation details are available on our project page.

\subsection{Hardware Experiments}
We deploy the trained RL policy on a Unitree G1 humanoid robot to validate our approach in real world.
The policy outputs joint position targets at $50$\,Hz, which are sent to the robot's low-level PD controller.
We test the policy on three types of granular terrain, namely basalt, a beach volleyball court, and a beach, as illustrated in Fig.~\ref{fig:hardware-results}.
As in simulation, we use success rate, velocity tracking error, and yaw tracking error as metrics.
We consider an episode a failure if the robot fails to maintain upright posture and requires human intervention.

\begin{figure}[t]
    \centering
    \begin{minipage}[t]{0.49\linewidth}
        \centering
        \includegraphics[width=\linewidth]{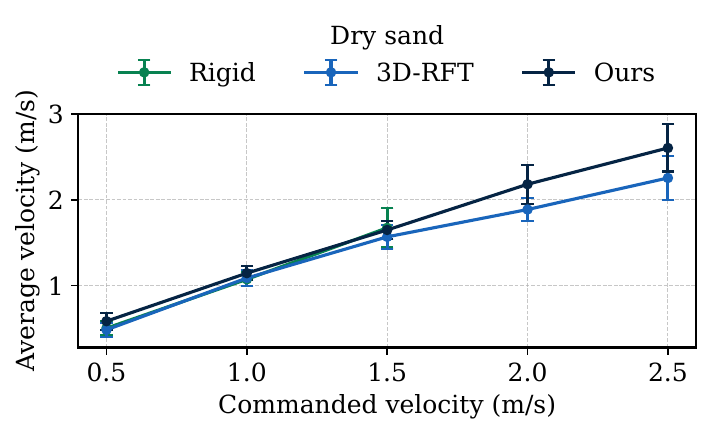}
    \end{minipage}\hfill
    \begin{minipage}[t]{0.49\linewidth}
        \centering
        \includegraphics[width=\linewidth]{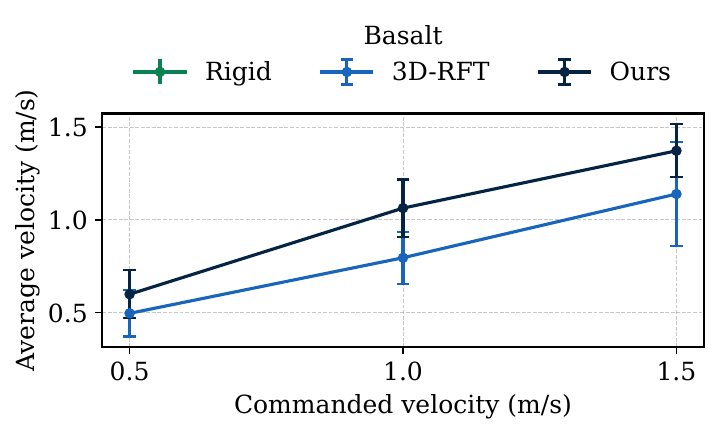}
    \end{minipage}
    \caption{
    Command velocity tracking of \modelrigid, \modelthreedrft, and \ours on dry sand (left) and basalt (right).
    Missing data points represent failed trials.
    }
    \label{fig:hardware-vel-tracking-v2}
    \vspace{-5mm}
\end{figure}

\textbf{3D RFT enables successful running behavior and our teacher-student policy improves velocity tracking performance.} 
We investigate the performance of the proposed controller on basalt and dry sand to analyze its robustness across different materials.
Specifically, we compare (1) \modelrigid, (2) \modelthreedrft, and (3) \ours, with commanded velocities of $0.5$, $1.0$, $1.5$, $2.0$, and $2.5$\,m/s as shown in Fig.~\ref{fig:hardware-vel-tracking-v2}.
On dry sand, all controllers show comparable velocity tracking up to $1.0$\,m/s.
At low speeds, the foot experiences small granular resistive forces, so \modelrigid can still maintain balance and achieve accurate velocity tracking.
However, at higher speeds ($\geq 1.5$\,m/s), \modelrigid experiences substantial degradation in performance due to increased perturbations to the foot, and fails to walk above $1.5$\,m/s. 
Furthermore, while \modelthreedrft and \ours maintain stable locomotion at the highest commanded velocity, \ours achieves the best velocity tracking.
On basalt, which is rougher and less uniform than the volleyball court sand, \modelrigid fails to maintain balance at all commanded velocities.
The combination of granular slippage and uneven terrain exceeds what the rigid-contact-trained policy can handle.
In contrast, \modelthreedrft and \ours successfully achieve locomotion across the entire range, with \ours achieving better velocity tracking than \modelthreedrft at all commanded velocities.
This highlights the advantage of our approach under varying terrain conditions.

\begin{figure}[t]
    \centering
    \includegraphics[width=1.0\linewidth]{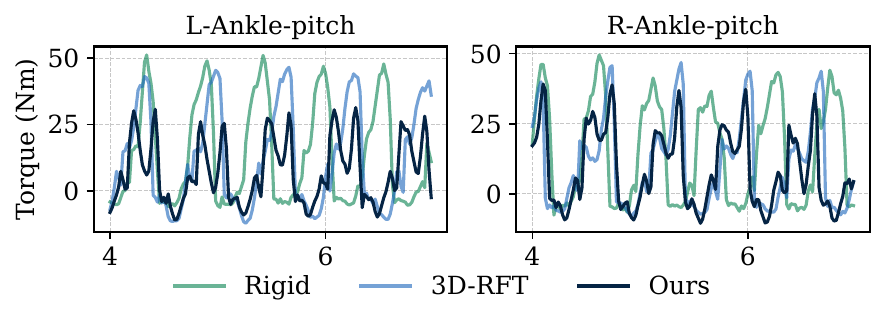}
    \caption{
    Estimated left and right ankle pitch torques during forward walking ($v = 1.5$\,m/s), comparing \modelrigid, \modelthreedrft, and \ours.
    }
    \label{fig:hardware-energy}
    \vspace{-3mm}
\end{figure}

\textbf{Our policy achieves compliant ankle control.}
Next, we analyze the resulting behavior of our teacher-student policy on hardware while walking on dry sand. 
To this end, we pay attention to the ankle pitch torque estimated from the measured motor currents, since it directly reflects how each controller absorbs terrain disturbance during stance.
Fig.~\ref{fig:hardware-energy} shows the estimated left and right ankle pitch torques during forward walking ($v = 1.5$\,m/s).
Our policy shows a lower peak ankle pitch torque than \modelrigid and \modelthreedrft on both legs, indicating compliant ankle control that minimizes terrain disturbance.
We conjecture that this is induced by the terrain encoder that estiamtes terrain stiffness, and baseline PPO without terrain estimation learns energy consuming gait that works for any terrain condition.
This finding aligns with the observation in the supplementary video that our controller exhibits the least stamping motion and kicks up the least sand among the three controllers.

\subsection{Locomotion under Terrain Transition}
\begin{figure}[t]
    \centering
    \includegraphics[width=1.0\linewidth]{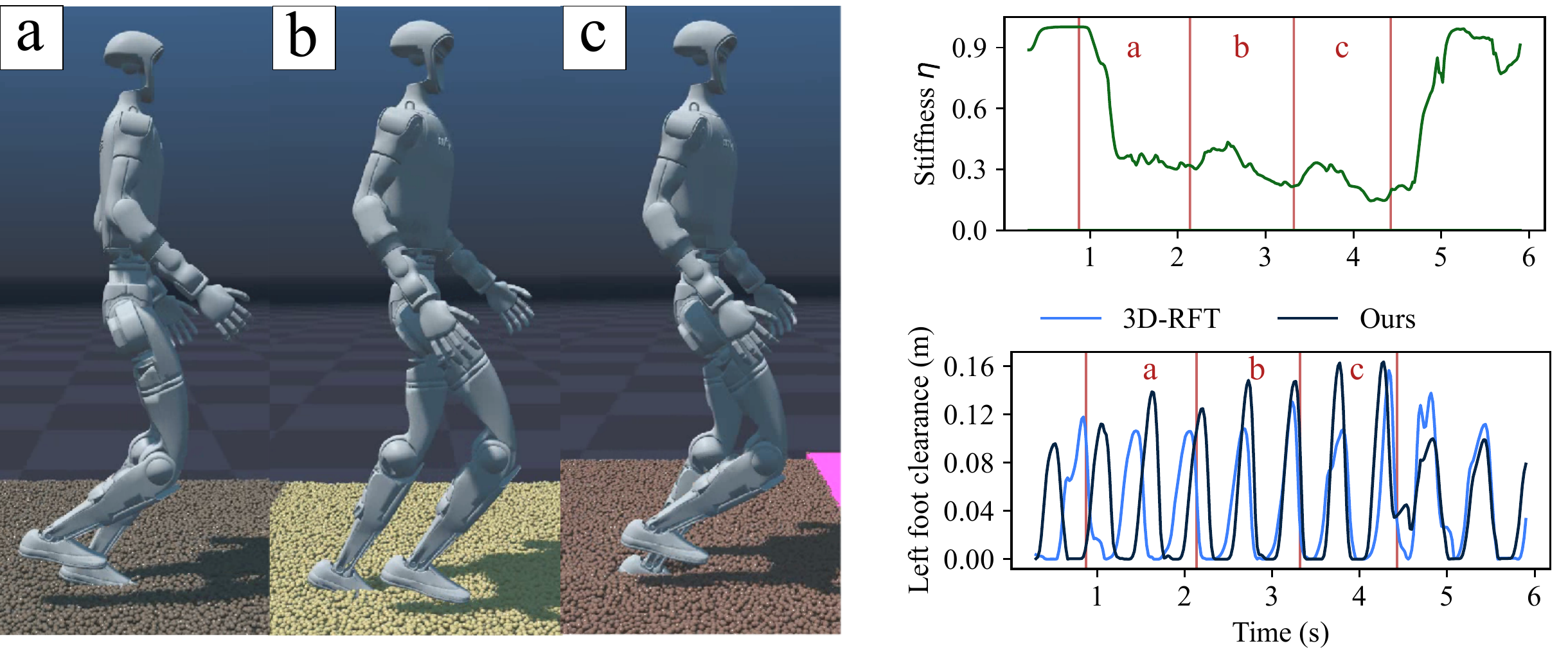}
    \caption{
        Online terrain estimation and adapted left foot clearance in MPM simulation.
        When the robot moves from stiff to softer granular layers, the predicted terrain stiffness drops, and foot clearance increases to clear the foot from the ground.
    }   
    \label{fig:adaptation}
    \vspace{-2mm}
\end{figure}

To evaluate terrain stiffness estimation and adaptation, we test our teacher-student policy in simulation on three consecutive sand layers of increasing softness (basalt, dry sand, and poppy seed) simulated with the MPM solver. 
On hardware, we achieve this terrain variation by building a pile of loosely packed sand.
In simulation, swing foot clearance stays near its nominal value of $9$\,cm on stiff terrain and rises up to $16$\,cm on softer terrain to avoid dragging the foot into the ground, as shown in Fig.~\ref{fig:adaptation}.
This adaptation tracks the decoded terrain stiffness $\mediascalingnorm$, which drops sharply as the robot enters softer sand layers, showing that the policy perceives terrain conditions and adjusts its behavior accordingly.

\begin{figure}[t]
    \centering
    \includegraphics[width=1.0\linewidth]
    {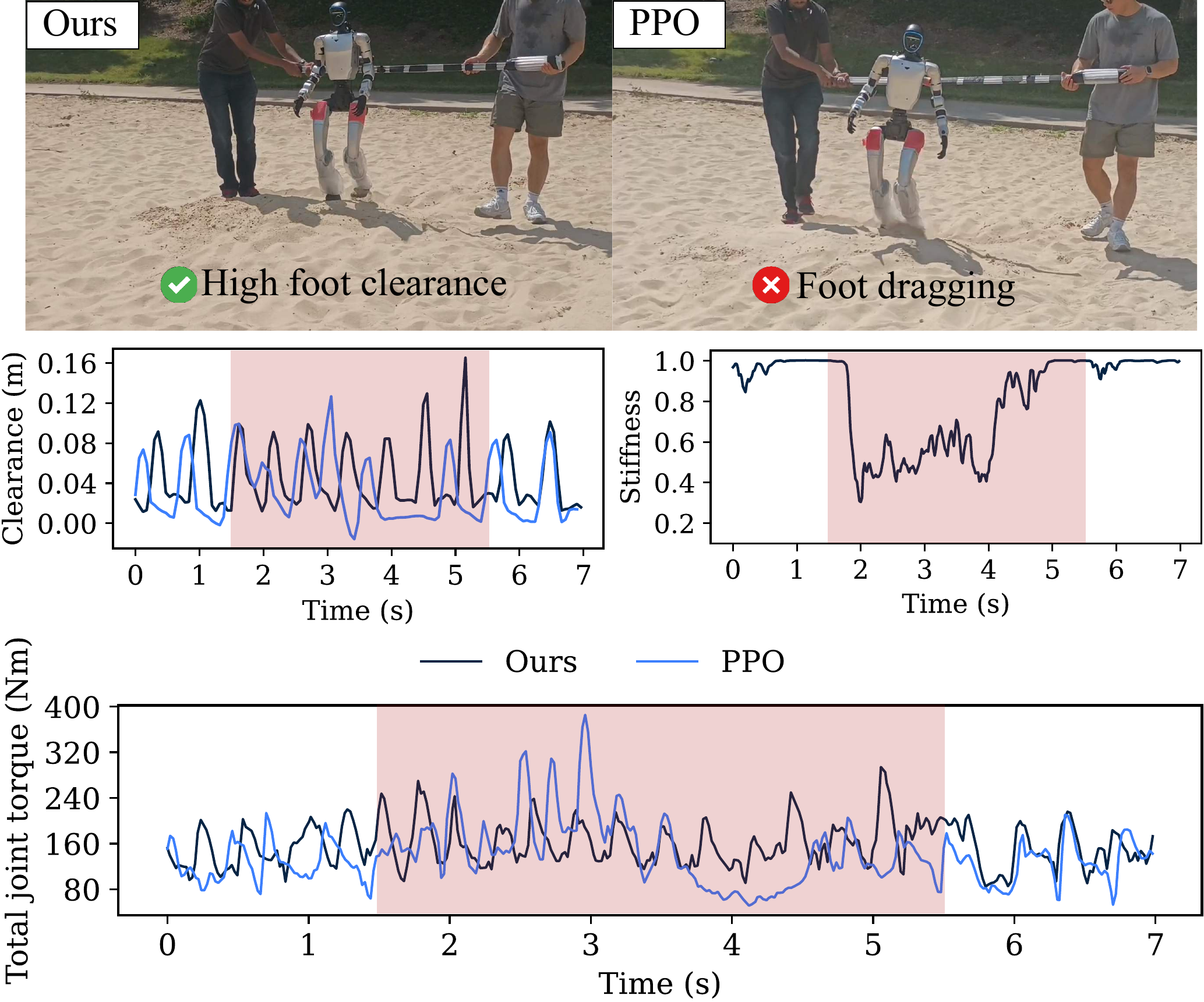}
    \caption{
        Online terrain stiffness estimation, swing foot clearance (left foot), and total joint torque (leg+waist) from hardware experiment.
        Our method maintains consistent walking behavior while baseline PPO loses balance during terrain transition.
        The shaded region marks the interval during which the foot is on the soft sand patch.
    }
    \label{fig:adaptation-hw}
    \vspace{-3mm}
\end{figure}

We observe the same trend on hardware while walking at $v = 0.5$\,m/s, as shown in Fig.~\ref{fig:adaptation-hw}.
\ours maintains positive foot clearance throughout the transition, while the PPO policy's foot clearance repeatedly drops below the ground reference, indicating that the foot drags into the sand rather than clearing it. This is corroborated by the recorded joint torques, where PPO exhibits large, reactive torque spikes indicative of repeated balance recovery, consistent with the foot-dragging observed in the hardware video, whereas \ours maintains a comparatively bounded torque profile. The predicted ground stiffness also tracks the terrain condition, indicating that our terrain-aware policy correctly identifies the terrain.

\section{Conclusion}
We present an RL framework for humanoid locomotion on granular terrain that combines a 3D RFT-based granular contact model with a terrain-adaptive teacher-student policy.
Simulation and hardware experiments demonstrate that explicit granular contact modeling during training is critical for robust locomotion, and that the learned terrain estimator enables adaptive behaviors such as swing foot clearance modulation on soft terrain.
Future work will (1) extend the proposed method to rough terrain such as slopes and wavy terrain and (2) study energy-efficient walking gaits on sand.

\section{ACKNOWLEDGMENTS}
We thank Cheng-Yuan Li, Sachin Kelkar, Jaehwi Jang, and Maaz Dossa for their help in the experiments. We also thank Shashank Agarwal and Deniz Kerimoglu for valuable discussions about RFT and MPM. We also thank Saaketh Reddy and Ziwon Yoon for providing us state estimation code.










\let\secfnt\undefined
\newfont{\secfnt}{ptmb8t at 10pt}

\bibliographystyle{IEEEtran}
\bibliography{./bibliography/reference}

\end{document}